\documentclass[runningheads, envcountsame, a4paper]{llncs}

\usepackage[ruled,vlined,linesnumbered]{algorithm2e}
\usepackage{amsmath}
\usepackage{amssymb}
\usepackage{cite}
\usepackage{epstopdf}
\usepackage[pdftex]{graphicx}
\usepackage[misc]{ifsym}
\usepackage{microtype}
\usepackage{multirow}
\usepackage{subcaption}
\usepackage[table,xcdraw]{xcolor}
\usepackage[hyphens]{url}

\usepackage{booktabs, amsfonts, dcolumn}
\newcolumntype{d}[1]{D..{#1}}

\begin{document}

\title{k-Winners-Take-All Ensemble Neural Network}

\titlerunning{k-Winners-Take-All Ensemble Neural Network}

\author{Abien Fred Agarap {\Letter} \and Arnulfo P. Azcarraga}

\authorrunning{A.F. Agarap and A. Azcarraga}

\institute{
College of Computer Studies \\
De La Salle University \\
2401 Taft Ave, Malate, Manila, 1004 Metro Manila, Philippines \\
\email{\{abien\_agarap, arnulfo.azcarraga\}}@dlsu.edu.ph\\
\url{https://dlsu.edu.ph}
}
\toctitle{k-Winners-Take-All Ensemble Neural Network}
\tocauthor{Abien~Fred~Agarap}
\tocauthor{Arnulfo~P.~Azcarraga}
\maketitle
\setcounter{footnote}{0}

\begin{abstract}
Ensembling is one approach that improves the performance of a neural network by combining a number of independent neural networks, usually by either averaging or summing up their individual outputs. We modify this ensembling approach by training the sub-networks concurrently instead of independently. This concurrent training of sub-networks leads them to cooperate with each other, and we refer to them as ``cooperative ensemble''. Meanwhile, the mixture-of-experts approach improves a neural network performance by dividing up a given dataset to its sub-networks. It then uses a gating network that assigns a specialization to each of its sub-networks called ``experts''. We improve on these aforementioned ways for combining a group of neural networks by using a k-Winners-Take-All (kWTA) activation function, that acts as the combination method for the outputs of each sub-network in the ensemble. We refer to this proposed model as ``kWTA ensemble neural networks'' (kWTA-ENN). With the kWTA activation function, the losing neurons of the sub-networks are inhibited while the winning neurons are retained. This results in sub-networks having some form of specialization but also sharing knowledge with one another. We compare our approach with the cooperative ensemble and mixture-of-experts, where we used a feed-forward neural network with one hidden layer having 100 neurons as the sub-network architecture. Our approach yields a better performance compared to the baseline models, reaching the following test accuracies on benchmark datasets: 98.34\% on MNIST, 88.06\% on Fashion-MNIST, 91.56\% on KMNIST, and 95.97\% on WDBC.
\end{abstract}

\section{Introduction and Related Works}
We use artificial neural networks in a myriad of automation tasks such as classification, regression, and translation among others. Neural networks would approach these tasks as a function approximation problem, wherein given a dataset of input-output pairs $D = \{(x_i, y_i) | x_i \in \mathcal{X}, y_i \in \mathcal{Y}\}$, their goal is to learn the mapping $\mathcal{F}: \mathcal{X} \mapsto \mathcal{Y}$. They accomplish this by optimizing their parameters $\theta$ with some modification mechanism, such as the retro-propagation of output errors\cite{rumelhart1986learning}. We then deem the parameters to be optimal if the neural network outputs are as close as possible to the target outputs in the training data, and if it can adequately generalize on previously unseen data. This can be achieved when a network is neither too simple (has a high bias) nor too complex (has a high variance).\\
\indent Through the years, combining a group of neural networks is among the simplest and most straightforward ways to achieve this feat. The two basic ways to combine neural networks are by ensembling\cite{breiman1996stacked, hansen1990neural, freund1996experiments, schapire1990strength}, and by using a mixture-of-experts (MoE)\cite{jacobs1991adaptive, jordan1992hierarchies}. In an ensemble, a group of independent neural networks is trained to learn the entire dataset. Meanwhile in MoE, each network is trained to learn their own and different subsets of the dataset.\\
\indent In this work, we use a group of neural networks for a classification task on the following benchmark datasets: MNIST \cite{lecun1998mnist}, Fashion-MNIST\cite{xiao2017fashion}, Kuzushiji-MNIST (KMNIST)\cite{clanuwat2018deep}, and Wisconsin Diagnostic Breast Cancer (WDBC)\cite{wolberg1992breast}. We introduce a variant of ensemble neural networks that uses a $k$-Winners-Take-All (kWTA) activation function to combine the outputs of its sub-networks instead of using averaging, summation, or voting schemes to combine such outputs. We then compare our approach with an MoE and a modified ensemble network on a classification task on the aforementioned datasets.

\subsection{Ensemble of Independent Networks}\label{sect:ensemble-learning}
We usually form an ensemble of networks by independently or sequentially (in the case of boosting) training them, and then by combining their outputs at test time usually by averaging\cite{breiman1996stacked} or voting\cite{hansen1990neural}. In this work, we opted to use the averaging scheme for ensembling.\\
\indent That is, we have a group of neural networks $f_1, \ldots, f_M$ parameterized by $\theta_1, \ldots, \theta_M$, and we compute its final output as,
\begin{align}\label{eq:averaging-ensemble}
    o = \frac{1}{M} \sum_{m=1}^{M} f_m(x; \theta_m)
\end{align}
\indent Each sub-network is trained independently to minimize their own loss function, e.g. cross entropy loss for classification, $\ell_{ce}(y, o) = -\sum y\log(o)$. Then Eq. \ref{eq:averaging-ensemble} is used to get the model outputs at test time.\\

\subsection{Mixture of Experts}
The Mixture-of-Experts (MoE) model consists of a set of $M$ ``expert'' neural networks $E_{1}, \ldots, E_{m}$ and a ``gating'' neural network $G$ \cite{jacobs1991adaptive}. The experts are assigned by the gating network to handle their own subset of the entire dataset. We compute the final output of this model using the following equation,
\begin{align}
    o = \sum \arg\max G(x) E_m(x)
\end{align}
where $G(x)$ is gating probability output to choose $E_{m}$ for a given input $x$. The gating network and the expert networks have their respective set of parameters.\\
Then, we compute the MoE model loss by using the following equation,
\begin{align}
    \mathcal{L}_{MoE}(x, y) = \frac{1}{M} \sum_{m = 1}^{M} \left[\frac{1}{n}\sum_{i = 1}^{n} \arg\max G(x_{i}) \cdot \ell_{ce}(y_i, E_{m}(x_i))\right]
\end{align}
where $\ell_{ce}$ is the cross entropy loss, $G(x)$ is the weighting factor to choose $E_m$.\\
\indent In this system, each expert learns to specialize on the cases where they perform well, and they are imposed to ignore the cases on which they do not perform well. With this learning paradigm, the experts become a function of a sub-region of the data space, and thus their set of learned weights highly differ from each other as opposed to traditional ensemble models that result to having almost identical weights for their learners.

\subsection{Cooperative Ensemble Learning}
We refer to the ensemble learning we described in Section \ref{sect:ensemble-learning} as traditional ensemble of independent neural networks. However, in our experiments, we trained the ensemble sub-networks concurrently instead of independently or sequentially. In Algorithm \ref{alg:cooperative-ensemble}, we present our modified version of the traditional ensemble, and we call it ``cooperative ensemble'' (CE) for the rest of this paper.\\
\begin{algorithm}[H]
\SetAlgoLined
\SetKwInOut{Input}{Input}
\SetKwInOut{Output}{Output}
\Input{Dataset $D = \{(x_i, y_i)|x_i \in \mathbb{R}^{d}, y_i = 1, \ldots, k\}$, randomly initialized networks $f_1, \ldots, f_M$ parameterized by $\theta_1, \ldots, \theta_M$}
\Output{Ensemble of $M$ trained networks $f_1, \ldots, f_M$}
 Initialization\;
 Sample mini-batch $B \subset D$\;
 \For{$t \gets 0$ \KwTo convergence}{
    \For {$m \gets 1$ \KwTo M}{
    \# Forward pass: Compute model outputs for mini-batch\\
    $\hat{y}_{m, 1},\ldots,\hat{y}_{m,B} = f_{m}(x_B)$\\
    }
    $o = \frac{1}{M} \sum_m^M \hat{y}_{m}$\\
    \# Backward pass: Update the models\\
    $\theta_m^{*} = \theta_m - \alpha \nabla \ell(y, o)$
 }
 \caption{Cooperative Ensemble Learning}
 \label{alg:cooperative-ensemble}
\end{algorithm}
\indent First, in a training loop, we compute each sub-network output $\hat{y}_{m, B} $ for mini-batches of data $B$ (line 6). Then, similar to a traditional ensemble, we compute the output of this model $o$ by averaging over the individual network outputs (line 8). Finally, we optimize the parameters of each sub-network in the ensemble based on the gradients of the loss between the ensemble network outputs $o$ and the target labels $y$ (line 10).\\
\indent In contrast, a traditional ensemble of independent networks train each sub-network independently before ensembling, thus not allowing an interaction among the members of the ensemble and not allowing a chance for each member to contribute to the knowledge of one another.\\
\indent Cooperative ensemble may have already been used in practice in the real world, but we take note of this variant for it presents itself as a more competitive baseline for our experimental model. This is because cooperative ensemble introduces some form of interaction among the sub-networks during training since there is an information feedback from the combination stage to the sub-network weights, thus giving each sub-network a chance to share their knowledge with one another\cite{liu1998cooperative}.\\
\indent The contributions of this study are as follows,
\begin{enumerate}
    \item The conceptual introduction of cooperative ensembling as a modification to the traditional ensemble of independent networks. The cooperative ensemble is a competitive baseline model for our experimental model (see Section \ref{sect:results-discussion}).
    \item We introduce an ensemble network that uses a kWTA activation function to combine its sub-network outputs (Section \ref{sect:methods}). Our approach presents better classification performance on the MNIST, Fashion-MNIST, KMNIST, and Wisconsin Diagnostic Breast Cancer (WDBC) datasets (see Section \ref{sect:results-discussion}).
\end{enumerate}
\section{Competitive Ensemble Learning}\label{sect:methods}
We take the cooperative ensembling approach further by introducing a competitive layer as a way to combine the outputs of the sub-networks in the ensemble.\\
\indent We propose to use a $k$-Winners-Take-All (kWTA) activation function for a fully connected layer which combines the sub-network outputs in the ensemble, and we call the resulting model ``kWTA ensemble neural network'' (kWTA-ENN). As per Majani et al. (1989) \cite{majani1989on}, the kWTA activation function admits $k \geq 1$ winners in a competition among neurons in a hidden layer of a neural network (see Eq. \ref{eq:kwta} for the kWTA activation function).
\begin{equation}\label{eq:kwta}
    \phi_k(z)_j =   \begin{cases}
                        z_j     &   z_j \in \{\max\limits_{k} z\} \\
                        0       &   z_j \not\in \{\max\limits_{k} z\}
                    \end{cases}
\end{equation}
where $z$ is an activation output, and $k$ is the percentage of winning neurons we want to get. We set $k=0.75$ in all our experiments, but it could still be optimized as it is a hyper-parameter. This kWTA activation function that we used is the classical one\cite{majani1989on} as we are only inhibiting the losing neurons in the competition while retaining the values of the winning neurons. Due to competition, the winning neurons gain the right to respond to particular subsets of the input data, as per Rumelhart \& Zipser (1985) \cite{rumelhart1985feature}.\\
\begin{algorithm}[!htb]
\SetAlgoLined
\SetKwInOut{Input}{Input}
\SetKwInOut{Output}{Output}
\Input{Dataset $D = \{(x_i, y_i)|x_i \in \mathbb{R}^{d}, y_i = 1, \ldots, k\}$, randomly initialized networks $f_1, \ldots, f_M$ parameterized by $\theta_1, \ldots, \theta_M$}
\Output{Ensemble of $M$ trained networks $f_1, \ldots, f_M$}
 Initialization\;
 Sample mini-batch $B \subset D$\;
 \For{$t \gets 0$ \KwTo convergence}{
    \For {$m \gets 1$ \KwTo M}{
    \# Forward pass: Compute model outputs for mini-batch\\
    $\hat{y}_{m, 1},\ldots,\hat{y}_{m,B} = f_{m}(x_B)$\\
    }
    $\hat{Y} = \hat{y}_{1, B},\ldots,\hat{y}_{M, B}$\\
    $z = \theta_z \hat{Y} + b_z$\\
    $o = \phi_k(z)$\\
    \# Backward pass: Update the models\\
    $\theta_m^{*} = \theta_m - \alpha \nabla \ell(y, o)$
 }
 \caption{k-Winners-Take-All Ensemble Network}
 \label{alg:kwta-enn}
\end{algorithm}
\indent We have seen the training algorithm for our cooperative ensemble in Algorithm \ref{alg:cooperative-ensemble}, wherein we train the sub-networks concurrently instead of independently or sequentially. We incorporate the same manner of training in kWTA-ENN, and we lay down our proposed training algorithm in Algorithm \ref{alg:kwta-enn}.\\
\indent Our model first computes the sub-network outputs $f_m(x_B)$ for each mini-batch of data $B$ (line 6) but as opposed to cooperative ensemble, we do not use a simple averaging of the sub-network outputs. Instead, we concatenate the sub-network outputs $\hat{Y}$ (line 8) and use it as an input to a fully connected layer (line 9). We then pass the fully connected layer output $z$ to the kWTA activation function (line 10). Finally, we update our ensemble based on the gradients of the loss between the kWTA-ENN outputs $o$ and the target labels $y$ (line 12).\\
\indent To further probe the effect of the kWTA activation function in the combination of sub-network outputs, we add a competition delay parameter $d$. We define this delay parameter as the number of initial training epochs where the kWTA activation function is not yet used on the fully connected layer output that combines the sub-network outputs. We set $d = 0; 3; 5; 7$.

\section{Experiments}\label{sect:results-discussion}
To demonstrate the performance gains using our approach, we used four benchmark datasets for evaluation: MNIST \cite{lecun1998mnist}, Fashion-MNIST\cite{xiao2017fashion}, KMNIST\cite{clanuwat2018deep}, and WDBC\cite{wolberg1992breast}. We ran each model ten times, and we report the average, best, and standard deviation of test accuracies for each of our model. Then, we ran a Kruskal-Wallis H test on the test accuracy results from ten runs of the baseline and experimental models.
\begin{table}[!htb]
    \centering
    \caption{Dataset statistics.}
    \begin{tabular}{c|c|c|c}
        \hline
        Dataset & \# Samples & Input Dimension & \# Classes \\
        \hline
        MNIST   &   70,000  &   784 & 10 \\
        Fashion-MNIST   &   70,000  &   784 & 10 \\
        KMNIST   &   70,000  &   784 & 10 \\
        WDBC    & 569   & 30    &   2 \\
        \hline
    \end{tabular}
    \label{tab:dataset_stats}
\end{table}
\subsection{Datasets Description}
We evaluate and compare our baseline and experimental models on three benchmark image datasets and one benchmark diagnostic dataset. We list the dataset statistics in Table \ref{tab:dataset_stats}.

All the *MNIST datasets consist of 60,000 training examples and 10,00 test examples each -- all in grayscale with $28\times28$ resolution. We flattened each image pixel matrix to a 784-dimensional vector.
\begin{itemize}
    \item \textbf{MNIST}. MNIST is a handwritten digit classification dataset\cite{lecun1998mnist}.
    \item \textbf{Fashion-MNIST}. Fashion-MNIST is said to be a more challenging alternative to MNIST that consists of fashion articles from Zalando\cite{xiao2017fashion}.
    \item \textbf{KMNIST}. Kuzushiji-MNIST (KMNIST) is another alternative to the MNIST dataset. Each of its classes represent one character representing each of the 10 rows of Hiragana\cite{clanuwat2018deep}.
    \item \textbf{WDBC}. The WDBC dataset is a binary classification dataset where its 30-dimensional features were computed from a digitized image of a fine needle aspirate of a breast mass\cite{wolberg1992breast}. It consists of 569 samples where 212 samples are malignant and 357 samples are benign. We randomly over-sampled the minority class in the dataset to account for its imbalanced class frequency distribution, thus increasing the number of samples to 714. We then splitted this dataset to 70\% training set and 30\% test set.
\end{itemize}
We randomly picked 10\% of the training samples for each of the dataset to serve as the validation dataset.
\subsection{Experimental Setup}
The code implementations for both our baseline and experimental models are found in \url{https://gitlab.com/afagarap/kwta-ensemble}.
\subsubsection{Hardware and Software Configuration}
We used a laptop computer with an Intel Core i5-6300HQ CPU with Nvidia GTX 960M GPU for training all our models. Then, we used the following arbitrarily chosen 10 random seeds for reproducibility: 42, 1234, 73, 1024, 86400, 31415, 2718, 30, 22, and 17. All our models were implemented in PyTorch 1.8.1 \cite{paszke2019pytorch} with some additional dependencies listed in the released source code.
\begin{table}[!htb]
\caption{Classification results on the benchmark datasets (bold values represent the best results) in terms of average, best, and standard deviation of test accuracies (in \%). Our kWTA-ENN achieves better test accuracies than our baseline models with statistical significance. * denotes at $p < 0.05$, $ns$ denotes not significant. }
\begin{subtable}[t]{0.5\textwidth}
\resizebox{\linewidth}{!}{%
\begin{tabular}{|c|c|c|c|c|c|c|c|}
\hline
\multicolumn{8}{|c|}{\textbf{MNIST}}                                                                                                                                                                 \\ \hline
\multirow{2}{*}{\textbf{\# nets}} & \multirow{2}{*}{\textbf{Acc}} & \multicolumn{4}{c|}{\textbf{kWTA-ENN}}                            & \multirow{2}{*}{\textbf{MoE}} & \multirow{2}{*}{\textbf{CE}} \\ \cline{3-6}
                                  &                               & \textbf{d = 0} & \textbf{d = 3} & \textbf{d = 5} & \textbf{d = 7} &                               &                              \\ \hline
\multirow{4}{*}{2}       & AVG                  & 98.16          & \textbf{98.18} & \textbf{98.18} & \textbf{98.18} & 96.43                         & 97.90                        \\ \cline{2-8} 
                                  & MAX                  & 98.28          & 98.28          & 98.28          & 98.28          & 96.66                         & 97.96                        \\ \cline{2-8} 
                                  & STD                  & 0.08           & 0.08           & 0.08           & 0.08           & 0.25                          & 0.05                         \\ \cline{2-8} 
                                  & \multicolumn{7}{c|}{* $H = 41.51,\ p = 7.39 \times 10^{-8}$}                                                                                                                     \\ \hline
\multirow{4}{*}{3}       & AVG                  & 98.24          & \textbf{98.26} & \textbf{98.26} & \textbf{98.26} & 94.67                         & 97.62                        \\ \cline{2-8} 
                                  & MAX                  & 98.36          & 98.39          & 98.39          & 98.39          & 96.33                         & 97.71                        \\ \cline{2-8} 
                                  & STD                  & 0.06           & 0.08           & 0.08           & 0.08           & 0.99                          & 0.05                         \\ \cline{2-8} 
                                  & \multicolumn{7}{c|}{* $H = 41.19,\ p = 8.61 \times 10^{-8}$}                                                                                                                     \\ \hline
\multirow{4}{*}{4}       & AVG                  & \textbf{98.30} & 98.27          & 98.27          & 98.27          & 92.349                        & 97.33                        \\ \cline{2-8} 
                                  & MAX                  & 98.43          & 98.39          & 98.39          & 98.39          & 95.02                         & 97.39                        \\ \cline{2-8} 
                                  & STD                  & 0.07           & 0.08           & 0.08           & 0.08           & 1.30                          & 0.05                         \\ \cline{2-8} 
                                  & \multicolumn{7}{c|}{* $H = 41.60,\ p = 7.11 \times 10^{-8}$}                                                                                                                     \\ \hline
\multirow{4}{*}{5}       & AVG                  & 98.33          & \textbf{98.34} & \textbf{98.34} & \textbf{98.34} & 90.63                         & 97.02                        \\ \cline{2-8} 
                                  & MAX                  & 98.52          & 98.42          & 98.42          & 98.42          & 91.94                         & 97.13                        \\ \cline{2-8} 
                                  & STD                  & 0.08           & 0.05           & 0.05           & 0.05           & 1.25                          & 0.06                         \\ \cline{2-8} 
                                  & \multicolumn{7}{c|}{* $H = 41.58,\ p = 7.17 \times 10^{-8}$}                                                                                                                     \\ \hline
\end{tabular}}
\label{tab:mnist-full-results}
\end{subtable}
\hspace{\fill}
\begin{subtable}[t]{0.5\textwidth}
\resizebox{\linewidth}{!}{%
\begin{tabular}{|c|c|c|c|c|c|c|c|}
\hline
\multicolumn{8}{|c|}{\textbf{Fashion-MNIST}}                                                                                                                                                         \\ \hline
\multirow{2}{*}{\textbf{\# nets}} & \multirow{2}{*}{\textbf{Acc}} & \multicolumn{4}{c|}{\textbf{kWTA-ENN}}                            & \multirow{2}{*}{\textbf{MoE}} & \multirow{2}{*}{\textbf{CE}} \\ \cline{3-6}
                                  &                               & \textbf{d = 0} & \textbf{d = 3} & \textbf{d = 5} & \textbf{d = 7} &                               &                              \\ \hline
\multirow{4}{*}{2}                & AVG                           & 87.53          & 87.54          & 87.54          & 87.54          & 86.59                         & \textbf{87.84}               \\ \cline{2-8} 
                                  & MAX                           & 87.78          & 87.70          & 87.70          & 87.70          & 87.54                         & 88.00                        \\ \cline{2-8} 
                                  & STD                           & 0.16           & 0.12           & 0.12           & 0.12           & 0.40                          & 0.11                         \\ \cline{2-8} 
                                  & \multicolumn{7}{c|}{* $H = 36.75,\ p = 6.72 \times 10^{-7}$}                                                                                                                      \\ \hline
\multirow{4}{*}{3}                & AVG                           & 87.73          & \textbf{87.81} & \textbf{87.81} & \textbf{87.81} & 85.54                         & 87.69                        \\ \cline{2-8} 
                                  & MAX                           & 88.01          & 88.10          & 88.10          & 88.10          & 87.15                         & 87.86                        \\ \cline{2-8} 
                                  & STD                           & 0.18           & 0.15           & 0.15           & 0.15           & 0.58                          & 0.09                         \\ \cline{2-8} 
                                  & \multicolumn{7}{c|}{* $H = 28.32,\ p = 3.16 \times 10^{-5}$}                                                                                                                      \\ \hline
\multirow{4}{*}{4}                & AVG                           & 87.88          & \textbf{87.93} & \textbf{87.93} & \textbf{87.93} & 84.47                         & 87.40                        \\ \cline{2-8} 
                                  & MAX                           & 88.22          & 88.15          & 88.15          & 88.15          & 86.69                         & 87.54                        \\ \cline{2-8} 
                                  & STD                           & 0.14           & 0.15           & 0.15           & 0.15           & 1.20                          & 0.09                         \\ \cline{2-8} 
                                  & \multicolumn{7}{c|}{* $H = 42.04,\ p = 5.78 \times 10^{-8}$}                                                                                                                      \\ \hline
\multirow{4}{*}{5}                & AVG                           & 87.99          & \textbf{88.06} & \textbf{88.06} & \textbf{88.06} & 82.89                         & 87.15                        \\ \cline{2-8} 
                                  & MAX                           & 88.22          & 88.27          & 88.27          & 88.27          & 85.80                         & 87.27                        \\ \cline{2-8} 
                                  & STD                           & 0.15           & 0.15           & 0.15           & 0.15           & 2.18                          & 0.05                         \\ \cline{2-8} 
                                  & \multicolumn{7}{c|}{* $H = 42.26,\ p = 5.22 \times 10^{-8}$}                                                                                                                      \\ \hline
\end{tabular}}
\label{tab:fmnist-full-results}
\end{subtable}
\bigskip
\begin{subtable}[t]{0.5\textwidth}
\resizebox{\linewidth}{!}{%
\begin{tabular}{|c|c|c|c|c|c|c|c|}
\hline
\multicolumn{8}{|c|}{\textbf{KMNIST}}                                                                                                                                                                \\ \hline
\multirow{2}{*}{\textbf{\# nets}} & \multirow{2}{*}{\textbf{Acc}} & \multicolumn{4}{c|}{\textbf{kWTA-ENN}}                            & \multirow{2}{*}{\textbf{MoE}} & \multirow{2}{*}{\textbf{CE}} \\ \cline{3-6}
                                  &                               & \textbf{d = 0} & \textbf{d = 3} & \textbf{d = 5} & \textbf{d = 7} &                               &                              \\ \hline
\multirow{4}{*}{2}                & AVG                           & \textbf{90.64} & 90.53          & 90.53          & 90.53          & 85.23                         & 89.94                        \\ \cline{2-8} 
                                  & MAX                           & 91.11          & 90.74          & 90.74          & 90.74          & 87.14                         & 90.19                        \\ \cline{2-8} 
                                  & STD                           & 0.29           & 0.12           & 0.12           & 0.12           & 0.99                          & 0.12                         \\ \cline{2-8} 
                                  & \multicolumn{7}{c|}{* $H = 41.63,\ p = 6.99 \times 10^{-8}$}                                                                                                                      \\ \hline
\multirow{4}{*}{3}                & AVG                           & 91.16          & \textbf{91.17} & \textbf{91.17} & \textbf{91.17} & 81.12                         & 89.47                        \\ \cline{2-8} 
                                  & MAX                           & 91.4           & 91.51          & 91.51          & 91.51          & 87.59                         & 89.61                        \\ \cline{2-8} 
                                  & STD                           & 0.14           & 0.19           & 0.19           & 0.19           & 2.77                          & 0.12                         \\ \cline{2-8} 
                                  & \multicolumn{7}{c|}{* $H = 41.09,\ p = 9.00 \times 10^{-8}$}                                                                                                                      \\ \hline
\multirow{4}{*}{4}                & AVG                           & \textbf{91.39} & 91.31          & 91.31          & 91.31          & 77.55                         & 88.72                        \\ \cline{2-8} 
                                  & MAX                           & 91.68          & 91.54          & 91.54          & 91.54          & 83.04                         & 88.94                        \\ \cline{2-8} 
                                  & STD                           & 0.18           & 0.15           & 0.15           & 0.15           & 2.89                          & 0.13                         \\ \cline{2-8} 
                                  & \multicolumn{7}{c|}{* $H = 41.67,\ p = 6.88 \times 10^{-8}$}                                                                                                                      \\ \hline
\multirow{4}{*}{5}                & AVG                           & \textbf{91.56} & 91.52          & 91.52          & 91.52          & 74.17                         & 87.87                        \\ \cline{2-8} 
                                  & MAX                           & 91.82          & 91.76          & 91.76          & 91.76          & 79.99                         & 88.02                        \\ \cline{2-8} 
                                  & STD                           & 0.16           & 0.18           & 0.18           & 0.18           & 3.47                          & 0.09                         \\ \cline{2-8} 
                                  & \multicolumn{7}{c|}{* $H = 41.28,\ p = 8.24 \times 10^{-8}$}                                                                                                                      \\ \hline
\end{tabular}}
\label{tab:kmnist-full-results}
\end{subtable}
\hspace{\fill}
\begin{subtable}[t]{0.5\textwidth}
\resizebox{\linewidth}{!}{%
\begin{tabular}{|c|c|c|c|c|c|c|c|}
\hline
\multicolumn{8}{|c|}{\textbf{WDBC}}                                                                                                                                                                  \\ \hline
\multirow{2}{*}{\textbf{\# nets}} & \multirow{2}{*}{\textbf{Acc}} & \multicolumn{4}{c|}{\textbf{kWTA-ENN}}                            & \multirow{2}{*}{\textbf{MoE}} & \multirow{2}{*}{\textbf{CE}} \\ \cline{3-6}
                                  &                               & \textbf{d = 0} & \textbf{d = 3} & \textbf{d = 5} & \textbf{d = 7} &                               &                              \\ \hline
\multirow{4}{*}{2}                & AVG                           & 95.43          & 95.36          & 95.36          & 95.36          & 94.49                         & \textbf{95.79}               \\ \cline{2-8} 
                                  & MAX                           & 98.62          & 98.62          & 98.62          & 98.62          & 98.57                         & 99.05                        \\ \cline{2-8} 
                                  & STD                           & 1.98           & 2.48           & 2.48           & 2.48           & 2.37                          & 2.13                         \\ \cline{2-8} 
                                  & \multicolumn{7}{c|}{(ns) $H = 1.40,\ p = 9.24 \times 10^{-1}$}                                                                                                                       \\ \hline
\multirow{4}{*}{3}                & AVG                           & 94.76          & \textbf{95.64} & \textbf{95.64} & \textbf{95.64} & 92.68                         & 95.35                        \\ \cline{2-8} 
                                  & MAX                           & 98.15          & 99.07          & 99.07          & 99.07          & 95.45                         & 98.17                        \\ \cline{2-8} 
                                  & STD                           & 1.92           & 2.33           & 2.33           & 2.33           & 2.36                          & 2.45                         \\ \cline{2-8} 
                                  & \multicolumn{7}{c|}{(ns) $H = 9.20,\ p = 1.02 \times 10^{-1}$}                                                                                                                       \\ \hline
\multirow{4}{*}{4}                & AVG                           & 94.98          & \textbf{95.97} & \textbf{95.97} & \textbf{95.97} & 91.79                         & 95.65                        \\ \cline{2-8} 
                                  & MAX                           & 98.62          & 98.62          & 98.62          & 98.62          & 96.67                         & 98.15                        \\ \cline{2-8} 
                                  & STD                           & 2.87           & 2.20           & 2.20           & 2.20           & 4.20                          & 2.00                         \\ \cline{2-8} 
                                  & \multicolumn{7}{c|}{* $H = 12.56,\ p = 2.78 \times 10^{-2}$}                                                                                                                      \\ \hline
\multirow{4}{*}{5}                & AVG                           & 95.03          & \textbf{95.40} & \textbf{95.40} & \textbf{95.40} & 90.93                         & 95.04                        \\ \cline{2-8} 
                                  & MAX                           & 98.61          & 99.05          & 99.05          & 99.05          & 96.33                         & 98.61                        \\ \cline{2-8} 
                                  & STD                           & 2.73           & 2.83           & 2.83           & 2.83           & 2.60                          & 2.47                         \\ \cline{2-8} 
                                  & \multicolumn{7}{c|}{* $H = 12.16,\ p = 3.27 \times 10^{-2}$}                                                                                                                      \\ \hline
\end{tabular}}
\label{tab:wdbc-full-results}
\end{subtable}
\label{tab:full-results}
\end{table}
\subsubsection{Training Details}
\indent For all our models, we used a feed-forward neural network with one hidden layer having 100 neurons as the sub-network, and then we vary the number of sub-networks per model from 2 to 5. The sub-network weights were initialized with Kaiming uniform initializer \cite{he2015delving}.\\
\indent We trained our baseline and experimental models on the MNIST, Fashion-MNIST, and KMNIST datasets using mini-batch stochastic gradient descent (SGD) with momentum\cite{qian1999momentum} of $9 \times 10^{-1}$, a learning rate of $1 \times 10^{-1}$ decaying to $1 \times 10^{-4}$, and weight decay of $1 \times 10^{-5}$ on a batch size of 100 for 10,800 iterations (equivalent to 20 epochs). As for the WDBC dataset, we used the same hyper-parameters except we trained our models for only 249 iterations (equivalent to 20 epochs).  All these hyper-parameters were arbitrarily chosen since we did not perform hyper-parameter tuning for any of our models. This makes the comparison fair for our baseline and experimental models, and we also did not have the computational resources to do so, which is why we chose a simple architecture as the sub-network.\\
\indent We recorded the accuracy and loss during both the training and validation phases. We then used the validation accuracy as the basis to checkpoint the best model parameters $\theta$ so far in the training. By the end of each training epoch, we load the best recorded parameters to be used by the model at test time.

\subsection{Classification Performance}
We evaluate the performance of our proposed approach in its different configurations as per the competition delay parameter $d$ and compare it with our baseline models: Mixture-of-Experts (MoE) and Cooperative Ensemble (CE). The empirical evidence shows our proposed approach outperforms our baseline models on the benchmark datasets we used. However, we are not able to observe a proper trend in performance with respect to the varying values of $d$, and thus it may warrant further investigation.\\
\indent For the full classification performance results of our baseline and experimental models, we refer the reader to Table \ref{tab:full-results}, from where we can observe the following:
\begin{enumerate}
    \item MoE performed the least among the models in our experiments, which may be justified with our choice of mini-batch size of 100. MoE performs better on larger datasets and/or larger batch sizes\cite{jacobs1991adaptive, shazeer2017outrageously}.
    \item CE is indeed a competitive baseline as we can see from the performance margins when compared to our proposed model.
    \item Our model in its different variations has consistently outperformed our baseline models in terms of average test accuracy (with the exception of two sub-networks for Fashion-MNIST and WDBC).
    \item Our model has higher margins on its improved test accuracy on the KMNIST dataset, which we find appealing since the said dataset is also supposed to be more difficult than the MNIST dataset and thus it better demonstrates the performance gains using our model.
    \item Finally, we can observe that there is a statistical significance among the differences in performance of the baseline and experimental models at $p < 0.05$ (on WDBC, for $M = 4, 5$ sub-networks), which indicates that the performance gains through our proposed approach are statistically significant. 
\end{enumerate}
\begin{figure*}[htb]
    \begin{subfigure}[b]{\linewidth}
    \includegraphics[width=\linewidth]{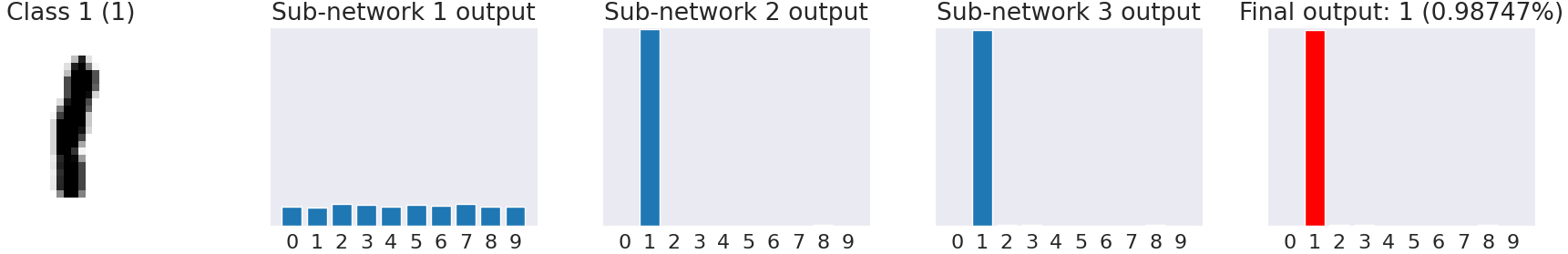}
    \caption{MoE}
    \label{fig:moe-mnist-per-class-logits}
    \end{subfigure}
    \begin{subfigure}[b]{\linewidth}
    \includegraphics[width=\linewidth]{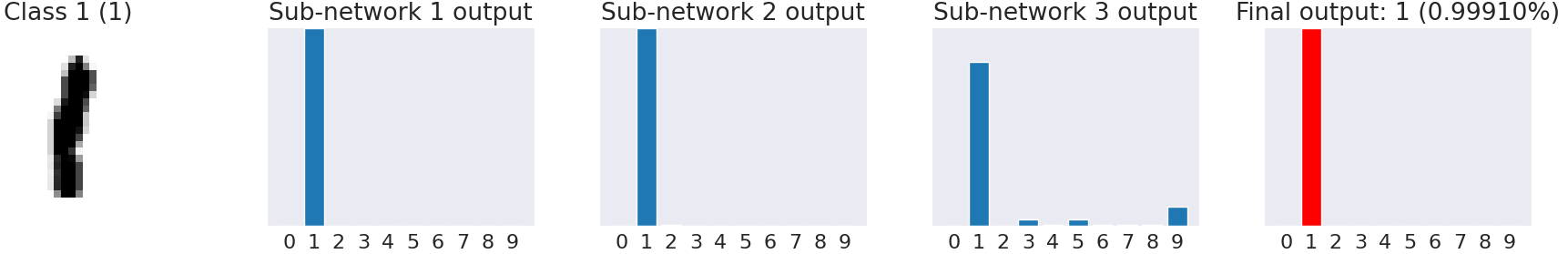}
    \caption{Cooperative Ensemble (CE)}
    \label{fig:ensemble-mnist-per-class-logits}
    \end{subfigure}
    \begin{subfigure}[b]{\linewidth}
    \includegraphics[width=\linewidth]{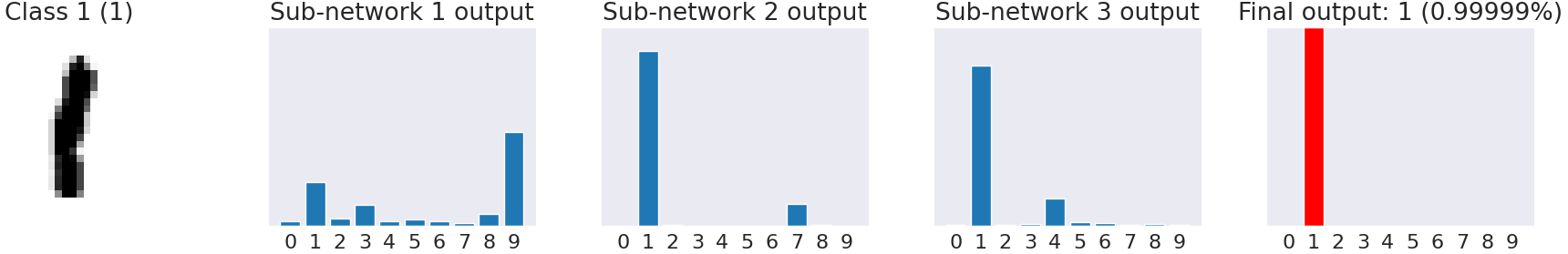}
    \caption{kWTA-ENN}
    \label{fig:kwta-mnist-per-class-logits}
    \end{subfigure}
    \caption{Predictions of each sub-network on a sample MNIST data and their respective final outputs. In \ref{fig:moe-mnist-per-class-logits}, we can infer that MoE sub-networks 2 and 3 are specializing on class 1. In \ref{fig:ensemble-mnist-per-class-logits}, all CE sub-networks have high probability outputs for class 1. In \ref{fig:kwta-mnist-per-class-logits}, all kWTA-ENN sub-networks contributed but with the kWTA activation function, the neurons for other classes were most likely inhibited at inference, thus its higher probability output than MoE and CE.}
    \label{fig:mnist-per-class-logits}
\end{figure*}
\begin{figure*}[!htb]
    \begin{subfigure}[htb]{\linewidth}
    \includegraphics[width=\linewidth]{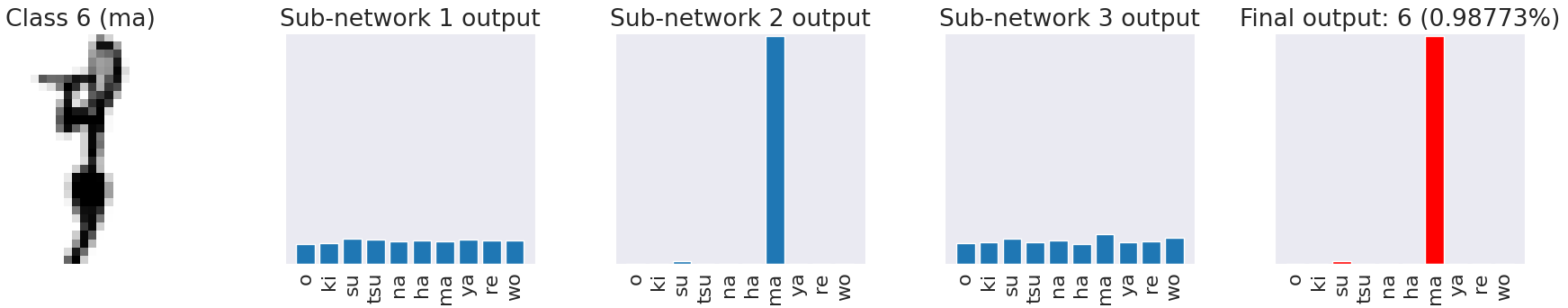}
    \caption{MoE}
    \label{fig:moe-kmnist-per-class-logits}
    \end{subfigure}
    \begin{subfigure}[b]{\linewidth}
    \includegraphics[width=\linewidth]{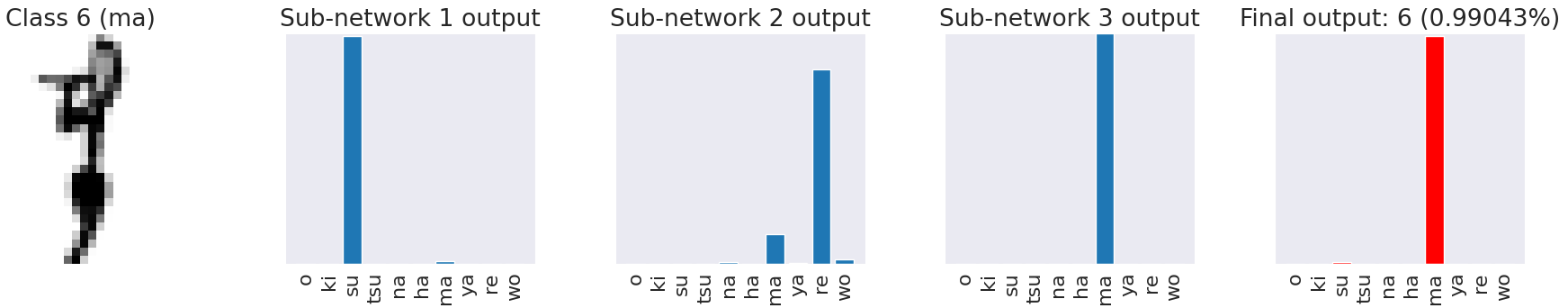}
    \caption{Cooperative Ensemble (CE)}
    \label{fig:ensemble-kmnist-per-class-logits}
    \end{subfigure}
    \begin{subfigure}[b]{\linewidth}
    \includegraphics[width=\linewidth]{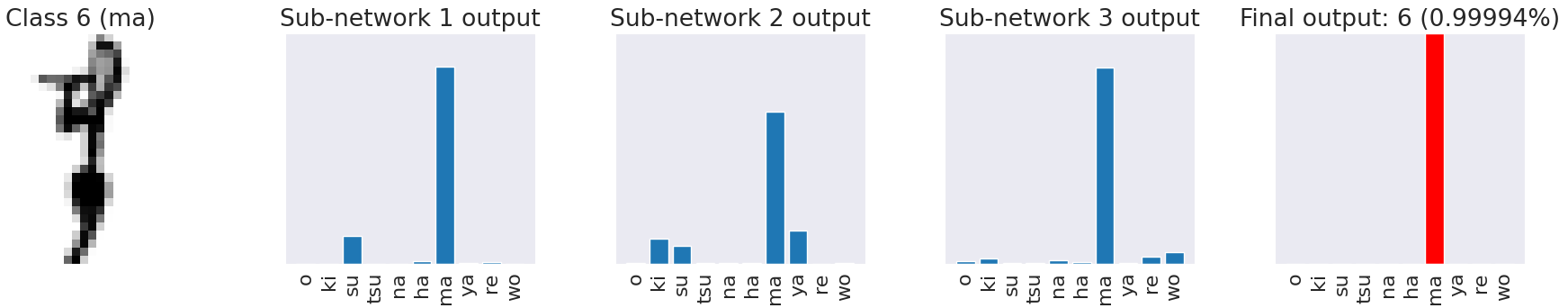}
    \caption{kWTA-ENN}
    \label{fig:kwta-kmnist-per-class-logits}
    \end{subfigure}
    \caption{Predictions of each sub-network on a sample KMNIST data and their respective final outputs. In \ref{fig:moe-kmnist-per-class-logits}, we can infer that MoE sub-network 2 is specializing on class 6 (``ma''). In \ref{fig:ensemble-kmnist-per-class-logits}, CE sub-network 3 was assisted by sub-network 2. In \ref{fig:kwta-kmnist-per-class-logits}, all kWTA-ENN sub-networks contributed but with the kWTA activation function, the neurons for other classes were most likely inhibited at inference, thus its higher probability output than MoE and CE.}
    \label{fig:kmnist-per-class-logits}
\end{figure*}
\subsection{Improving cooperation through competitive learning}
In the context of our work, we refer to \textit{cooperation} in a group of neural networks as the phenomenon when the members of the group contribute to the overall group performance. For instance, in CE, all the sub-networks contribute to the knowledge of one another as opposed to the traditional ensemble, where there is no interaction among the ensemble members\cite{liu1998cooperative}. Meanwhile, \textit{specialization} is when members of a group of neural networks are tasked to a specific subset of the input data, which is the intention behind the design of MoE\cite{jacobs1991adaptive}. In this respect, \textit{competition} can be thought of leading to specialization since it is when the winning units gain the right to respond to a particular subset of the dataset. We argue that with our proposed approach, we employ the notion of all three: competition, specialization, and cooperation.\\
\begin{table}[htb!]
\caption{Classification results of each kWTA-ENN sub-network and kWTA-ENN itself on MNIST (\ref{tab:per-class-mnist}) and KMNIST (\ref{tab:per-class-kmnist}) datasets. The tables show the test accuracy of each sub-network on each dataset class, indicating a degree of specialization among the sub-networks. Furthermore, the final model accuracy on each class shows that combining the sub-network outputs have stronger predictive capability. These divisions were in no way pre-determined but they show how cooperation by specialization can be done through competitive ensemble.}
\begin{subtable}[t]{\textwidth}
\caption{}
\resizebox{\linewidth}{!}{%
\begin{tabular}{ccccllccccclccccclcccccc}
\multicolumn{1}{l}{} &
  \multicolumn{3}{c}{n = 2} &
   &
   &
  \multicolumn{4}{c}{n = 3} &
  \multicolumn{1}{l}{} &
   &
  \multicolumn{5}{c}{n = 4} &
   &
  \multicolumn{6}{c}{n = 5} \\ \cline{2-4} \cline{7-10} \cline{13-17} \cline{19-24} 
\multicolumn{1}{c|}{0} &
  \multicolumn{1}{c|}{\cellcolor[HTML]{6CC499}93.65} &
  \multicolumn{1}{c|}{\cellcolor[HTML]{6FC59B}93.07} &
  \multicolumn{1}{c|}{\cellcolor[HTML]{5ABD8C}98.38} &
   &
  \multicolumn{1}{l|}{} &
  \multicolumn{1}{c|}{\cellcolor[HTML]{60BF91}96.44} &
  \multicolumn{1}{c|}{\cellcolor[HTML]{73C79E}90.38} &
  \multicolumn{1}{c|}{\cellcolor[HTML]{66C295}94.49} &
  \multicolumn{1}{c|}{\cellcolor[HTML]{59BC8C}98.78} &
   &
  \multicolumn{1}{l|}{} &
  \multicolumn{1}{c|}{\cellcolor[HTML]{9FD8BC}71.55} &
  \multicolumn{1}{c|}{\cellcolor[HTML]{7FCBA6}84.36} &
  \multicolumn{1}{c|}{\cellcolor[HTML]{59BC8B}99.56} &
  \multicolumn{1}{c|}{\cellcolor[HTML]{A5DBC0}69.15} &
  \multicolumn{1}{c|}{\cellcolor[HTML]{5BBD8D}98.68} &
  \multicolumn{1}{l|}{} &
  \multicolumn{1}{c|}{\cellcolor[HTML]{5FBE90}96.10} &
  \multicolumn{1}{c|}{\cellcolor[HTML]{65C194}93.11} &
  \multicolumn{1}{c|}{\cellcolor[HTML]{76C8A0}85.83} &
  \multicolumn{1}{c|}{\cellcolor[HTML]{61C091}94.91} &
  \multicolumn{1}{c|}{\cellcolor[HTML]{8ED2B1}75.07} &
  \multicolumn{1}{c|}{\cellcolor[HTML]{57BB8A}99.29} \\ \cline{2-4} \cline{7-10} \cline{13-17} \cline{19-24} 
\multicolumn{1}{c|}{1} &
  \multicolumn{1}{c|}{\cellcolor[HTML]{63C093}95.97} &
  \multicolumn{1}{c|}{\cellcolor[HTML]{8CD1AF}85.42} &
  \multicolumn{1}{c|}{\cellcolor[HTML]{57BB8A}99.03} &
   &
  \multicolumn{1}{l|}{} &
  \multicolumn{1}{c|}{\cellcolor[HTML]{C0E6D4}65.52} &
  \multicolumn{1}{c|}{\cellcolor[HTML]{61BF91}96.13} &
  \multicolumn{1}{c|}{\cellcolor[HTML]{A0D9BD}75.91} &
  \multicolumn{1}{c|}{\cellcolor[HTML]{57BB8A}99.30} &
   &
  \multicolumn{1}{l|}{} &
  \multicolumn{1}{c|}{\cellcolor[HTML]{C9EADA}54.44} &
  \multicolumn{1}{c|}{\cellcolor[HTML]{6EC59A}91.07} &
  \multicolumn{1}{c|}{\cellcolor[HTML]{57BB8A}100.00} &
  \multicolumn{1}{c|}{\cellcolor[HTML]{CEECDD}52.46} &
  \multicolumn{1}{c|}{\cellcolor[HTML]{59BC8B}99.47} &
  \multicolumn{1}{l|}{} &
  \multicolumn{1}{c|}{\cellcolor[HTML]{71C69C}87.96} &
  \multicolumn{1}{c|}{\cellcolor[HTML]{97D5B7}71.03} &
  \multicolumn{1}{c|}{\cellcolor[HTML]{DBF1E6}NaN} &
  \multicolumn{1}{c|}{\cellcolor[HTML]{B4E1CB}58.35} &
  \multicolumn{1}{c|}{\cellcolor[HTML]{96D5B6}71.60} &
  \multicolumn{1}{c|}{\cellcolor[HTML]{58BC8B}99.03} \\ \cline{2-4} \cline{7-10} \cline{13-17} \cline{19-24} 
\multicolumn{1}{c|}{2} &
  \multicolumn{1}{c|}{\cellcolor[HTML]{91D3B3}84.15} &
  \multicolumn{1}{c|}{\cellcolor[HTML]{85CEAA}87.29} &
  \multicolumn{1}{c|}{\cellcolor[HTML]{59BC8C}98.55} &
   &
  \multicolumn{1}{l|}{} &
  \multicolumn{1}{c|}{\cellcolor[HTML]{C8E9D9}62.92} &
  \multicolumn{1}{c|}{\cellcolor[HTML]{65C194}94.89} &
  \multicolumn{1}{c|}{\cellcolor[HTML]{8AD0AE}83.01} &
  \multicolumn{1}{c|}{\cellcolor[HTML]{5BBD8D}98.26} &
   &
  \multicolumn{1}{l|}{} &
  \multicolumn{1}{c|}{\cellcolor[HTML]{82CDA8}82.80} &
  \multicolumn{1}{c|}{\cellcolor[HTML]{97D5B7}74.53} &
  \multicolumn{1}{c|}{\cellcolor[HTML]{F8FCFA}35.96} &
  \multicolumn{1}{c|}{\cellcolor[HTML]{90D2B2}77.51} &
  \multicolumn{1}{c|}{\cellcolor[HTML]{5DBE8E}97.88} &
  \multicolumn{1}{l|}{} &
  \multicolumn{1}{c|}{\cellcolor[HTML]{8DD1B0}75.66} &
  \multicolumn{1}{c|}{\cellcolor[HTML]{B8E3CE}56.49} &
  \multicolumn{1}{c|}{\cellcolor[HTML]{E4F4EC}37.32} &
  \multicolumn{1}{c|}{\cellcolor[HTML]{82CDA8}80.63} &
  \multicolumn{1}{c|}{\cellcolor[HTML]{5FBF90}95.86} &
  \multicolumn{1}{c|}{\cellcolor[HTML]{59BC8C}98.44} \\ \cline{2-4} \cline{7-10} \cline{13-17} \cline{19-24} 
\multicolumn{1}{c|}{3} &
  \multicolumn{1}{c|}{\cellcolor[HTML]{9ED8BB}80.95} &
  \multicolumn{1}{c|}{\cellcolor[HTML]{FFFFFF}55.70} &
  \multicolumn{1}{c|}{\cellcolor[HTML]{5DBE8E}97.64} &
   &
  \multicolumn{1}{l|}{} &
  \multicolumn{1}{c|}{\cellcolor[HTML]{FFFFFF}45.22} &
  \multicolumn{1}{c|}{\cellcolor[HTML]{71C69C}90.98} &
  \multicolumn{1}{c|}{\cellcolor[HTML]{71C69C}91.16} &
  \multicolumn{1}{c|}{\cellcolor[HTML]{5ABD8C}98.42} &
   &
  \multicolumn{1}{l|}{} &
  \multicolumn{1}{c|}{\cellcolor[HTML]{F5FBF8}36.96} &
  \multicolumn{1}{c|}{\cellcolor[HTML]{B3E1CA}63.28} &
  \multicolumn{1}{c|}{\cellcolor[HTML]{A1D9BE}70.62} &
  \multicolumn{1}{c|}{\cellcolor[HTML]{B4E1CB}62.89} &
  \multicolumn{1}{c|}{\cellcolor[HTML]{5CBD8D}98.32} &
  \multicolumn{1}{l|}{} &
  \multicolumn{1}{c|}{\cellcolor[HTML]{E1F3EA}38.42} &
  \multicolumn{1}{c|}{\cellcolor[HTML]{78C9A1}84.77} &
  \multicolumn{1}{c|}{\cellcolor[HTML]{E2F4EB}37.93} &
  \multicolumn{1}{c|}{\cellcolor[HTML]{D4EEE1}44.11} &
  \multicolumn{1}{c|}{\cellcolor[HTML]{7ECBA5}82.18} &
  \multicolumn{1}{c|}{\cellcolor[HTML]{5BBD8D}97.85} \\ \cline{2-4} \cline{7-10} \cline{13-17} \cline{19-24} 
\multicolumn{1}{c|}{4} &
  \multicolumn{1}{c|}{\cellcolor[HTML]{86CEAB}87.01} &
  \multicolumn{1}{c|}{\cellcolor[HTML]{81CCA8}88.22} &
  \multicolumn{1}{c|}{\cellcolor[HTML]{5ABC8C}98.47} &
   &
  \multicolumn{1}{l|}{} &
  \multicolumn{1}{c|}{\cellcolor[HTML]{E7F5EE}53.23} &
  \multicolumn{1}{c|}{\cellcolor[HTML]{7CCAA4}87.55} &
  \multicolumn{1}{c|}{\cellcolor[HTML]{AADDC4}72.75} &
  \multicolumn{1}{c|}{\cellcolor[HTML]{5BBD8D}98.27} &
   &
  \multicolumn{1}{l|}{} &
  \multicolumn{1}{c|}{\cellcolor[HTML]{92D3B3}76.58} &
  \multicolumn{1}{c|}{\cellcolor[HTML]{E7F6EE}42.58} &
  \multicolumn{1}{c|}{\cellcolor[HTML]{A5DBC1}68.86} &
  \multicolumn{1}{c|}{\cellcolor[HTML]{88CFAC}80.73} &
  \multicolumn{1}{c|}{\cellcolor[HTML]{5CBD8E}98.07} &
  \multicolumn{1}{l|}{} &
  \multicolumn{1}{c|}{\cellcolor[HTML]{C0E6D3}53.01} &
  \multicolumn{1}{c|}{\cellcolor[HTML]{80CCA7}81.43} &
  \multicolumn{1}{c|}{\cellcolor[HTML]{ABDDC4}62.47} &
  \multicolumn{1}{c|}{\cellcolor[HTML]{7DCBA5}82.61} &
  \multicolumn{1}{c|}{\cellcolor[HTML]{C3E7D5}51.95} &
  \multicolumn{1}{c|}{\cellcolor[HTML]{5ABD8D}97.97} \\ \cline{2-4} \cline{7-10} \cline{13-17} \cline{19-24} 
\multicolumn{1}{c|}{5} &
  \multicolumn{1}{c|}{\cellcolor[HTML]{82CDA8}88.11} &
  \multicolumn{1}{c|}{\cellcolor[HTML]{8BD0AE}85.79} &
  \multicolumn{1}{c|}{\cellcolor[HTML]{58BC8B}98.87} &
   &
  \multicolumn{1}{l|}{} &
  \multicolumn{1}{c|}{\cellcolor[HTML]{6CC499}92.61} &
  \multicolumn{1}{c|}{\cellcolor[HTML]{B5E1CB}69.26} &
  \multicolumn{1}{c|}{\cellcolor[HTML]{BBE4D0}67.32} &
  \multicolumn{1}{c|}{\cellcolor[HTML]{59BC8C}98.65} &
   &
  \multicolumn{1}{l|}{} &
  \multicolumn{1}{c|}{\cellcolor[HTML]{BCE4D0}59.88} &
  \multicolumn{1}{c|}{\cellcolor[HTML]{C4E7D6}56.59} &
  \multicolumn{1}{c|}{\cellcolor[HTML]{A6DBC1}68.49} &
  \multicolumn{1}{c|}{\cellcolor[HTML]{A2DABE}70.17} &
  \multicolumn{1}{c|}{\cellcolor[HTML]{5ABD8C}98.87} &
  \multicolumn{1}{l|}{} &
  \multicolumn{1}{c|}{\cellcolor[HTML]{DBF1E6}40.99} &
  \multicolumn{1}{c|}{\cellcolor[HTML]{86CEAB}78.57} &
  \multicolumn{1}{c|}{\cellcolor[HTML]{C5E8D7}50.75} &
  \multicolumn{1}{c|}{\cellcolor[HTML]{FFFFFF}25.00} &
  \multicolumn{1}{c|}{\cellcolor[HTML]{BFE5D3}53.43} &
  \multicolumn{1}{c|}{\cellcolor[HTML]{58BC8B}99.20} \\ \cline{2-4} \cline{7-10} \cline{13-17} \cline{19-24} 
\multicolumn{1}{c|}{6} &
  \multicolumn{1}{c|}{\cellcolor[HTML]{5DBE8F}97.50} &
  \multicolumn{1}{c|}{\cellcolor[HTML]{6EC59A}93.12} &
  \multicolumn{1}{c|}{\cellcolor[HTML]{5ABC8C}98.43} &
   &
  \multicolumn{1}{l|}{} &
  \multicolumn{1}{c|}{\cellcolor[HTML]{89D0AD}83.32} &
  \multicolumn{1}{c|}{\cellcolor[HTML]{69C296}93.76} &
  \multicolumn{1}{c|}{\cellcolor[HTML]{74C79E}90.21} &
  \multicolumn{1}{c|}{\cellcolor[HTML]{59BC8B}98.85} &
   &
  \multicolumn{1}{l|}{} &
  \multicolumn{1}{c|}{\cellcolor[HTML]{6CC499}91.64} &
  \multicolumn{1}{c|}{\cellcolor[HTML]{D4EEE2}50.00} &
  \multicolumn{1}{c|}{\cellcolor[HTML]{8DD1B0}78.41} &
  \multicolumn{1}{c|}{\cellcolor[HTML]{84CDA9}82.21} &
  \multicolumn{1}{c|}{\cellcolor[HTML]{5BBD8D}98.53} &
  \multicolumn{1}{l|}{} &
  \multicolumn{1}{c|}{\cellcolor[HTML]{79C9A2}84.51} &
  \multicolumn{1}{c|}{\cellcolor[HTML]{8BD0AE}76.62} &
  \multicolumn{1}{c|}{\cellcolor[HTML]{6BC398}90.84} &
  \multicolumn{1}{c|}{\cellcolor[HTML]{64C193}93.66} &
  \multicolumn{1}{c|}{\cellcolor[HTML]{8FD2B1}74.83} &
  \multicolumn{1}{c|}{\cellcolor[HTML]{59BC8C}98.54} \\ \cline{2-4} \cline{7-10} \cline{13-17} \cline{19-24} 
\multicolumn{1}{c|}{7} &
  \multicolumn{1}{c|}{\cellcolor[HTML]{B7E2CD}74.34} &
  \multicolumn{1}{c|}{\cellcolor[HTML]{96D5B6}82.96} &
  \multicolumn{1}{c|}{\cellcolor[HTML]{5BBD8D}98.06} &
   &
  \multicolumn{1}{l|}{} &
  \multicolumn{1}{c|}{\cellcolor[HTML]{84CDA9}85.11} &
  \multicolumn{1}{c|}{\cellcolor[HTML]{A5DBC1}74.32} &
  \multicolumn{1}{c|}{\cellcolor[HTML]{72C69D}90.80} &
  \multicolumn{1}{c|}{\cellcolor[HTML]{5BBD8D}98.25} &
   &
  \multicolumn{1}{l|}{} &
  \multicolumn{1}{c|}{\cellcolor[HTML]{B9E3CF}60.91} &
  \multicolumn{1}{c|}{\cellcolor[HTML]{9DD7BB}72.37} &
  \multicolumn{1}{c|}{\cellcolor[HTML]{C3E7D5}57.07} &
  \multicolumn{1}{c|}{\cellcolor[HTML]{D7EFE3}48.87} &
  \multicolumn{1}{c|}{\cellcolor[HTML]{5BBD8D}98.64} &
  \multicolumn{1}{l|}{} &
  \multicolumn{1}{c|}{\cellcolor[HTML]{8BD1AF}76.30} &
  \multicolumn{1}{c|}{\cellcolor[HTML]{88CFAC}78.04} &
  \multicolumn{1}{c|}{\cellcolor[HTML]{8CD1AF}76.23} &
  \multicolumn{1}{c|}{\cellcolor[HTML]{93D4B4}72.87} &
  \multicolumn{1}{c|}{\cellcolor[HTML]{C6E8D7}50.41} &
  \multicolumn{1}{c|}{\cellcolor[HTML]{59BC8B}98.73} \\ \cline{2-4} \cline{7-10} \cline{13-17} \cline{19-24} 
\multicolumn{1}{c|}{8} &
  \multicolumn{1}{c|}{\cellcolor[HTML]{C7E9D8}70.23} &
  \multicolumn{1}{c|}{\cellcolor[HTML]{6DC499}93.52} &
  \multicolumn{1}{c|}{\cellcolor[HTML]{5EBE8F}97.44} &
   &
  \multicolumn{1}{l|}{} &
  \multicolumn{1}{c|}{\cellcolor[HTML]{D9F0E5}57.64} &
  \multicolumn{1}{c|}{\cellcolor[HTML]{9CD7BA}77.35} &
  \multicolumn{1}{c|}{\cellcolor[HTML]{AADDC4}72.58} &
  \multicolumn{1}{c|}{\cellcolor[HTML]{5DBE8F}97.44} &
   &
  \multicolumn{1}{l|}{} &
  \multicolumn{1}{c|}{\cellcolor[HTML]{A2DABE}70.26} &
  \multicolumn{1}{c|}{\cellcolor[HTML]{D3EDE0}50.69} &
  \multicolumn{1}{c|}{\cellcolor[HTML]{D0ECDF}51.66} &
  \multicolumn{1}{c|}{\cellcolor[HTML]{C8E9D9}54.81} &
  \multicolumn{1}{c|}{\cellcolor[HTML]{5DBE8F}97.64} &
  \multicolumn{1}{l|}{} &
  \multicolumn{1}{c|}{\cellcolor[HTML]{F3FBF7}30.41} &
  \multicolumn{1}{c|}{\cellcolor[HTML]{E9F7F0}34.74} &
  \multicolumn{1}{c|}{\cellcolor[HTML]{AADDC4}62.99} &
  \multicolumn{1}{c|}{\cellcolor[HTML]{ECF8F2}33.42} &
  \multicolumn{1}{c|}{\cellcolor[HTML]{91D3B2}74.02} &
  \multicolumn{1}{c|}{\cellcolor[HTML]{5BBD8D}97.64} \\ \cline{2-4} \cline{7-10} \cline{13-17} \cline{19-24} 
\multicolumn{1}{c|}{9} &
  \multicolumn{1}{c|}{\cellcolor[HTML]{65C194}95.61} &
  \multicolumn{1}{c|}{\cellcolor[HTML]{A1D9BE}79.95} &
  \multicolumn{1}{c|}{\cellcolor[HTML]{5CBD8E}97.80} &
   &
  \multicolumn{1}{l|}{} &
  \multicolumn{1}{c|}{\cellcolor[HTML]{EBF7F1}51.96} &
  \multicolumn{1}{c|}{\cellcolor[HTML]{AADDC4}72.86} &
  \multicolumn{1}{c|}{\cellcolor[HTML]{FFFFFF}45.49} &
  \multicolumn{1}{c|}{\cellcolor[HTML]{5DBE8F}97.42} &
   &
  \multicolumn{1}{l|}{} &
  \multicolumn{1}{c|}{\cellcolor[HTML]{BEE5D2}59.16} &
  \multicolumn{1}{c|}{\cellcolor[HTML]{FFFFFF}32.77} &
  \multicolumn{1}{c|}{\cellcolor[HTML]{86CEAB}81.57} &
  \multicolumn{1}{c|}{\cellcolor[HTML]{97D5B7}74.63} &
  \multicolumn{1}{c|}{\cellcolor[HTML]{5CBE8E}98.01} &
  \multicolumn{1}{l|}{} &
  \multicolumn{1}{c|}{\cellcolor[HTML]{C9E9D9}49.32} &
  \multicolumn{1}{c|}{\cellcolor[HTML]{6FC59B}88.89} &
  \multicolumn{1}{c|}{\cellcolor[HTML]{E9F6F0}34.84} &
  \multicolumn{1}{c|}{\cellcolor[HTML]{B1E0C9}59.78} &
  \multicolumn{1}{c|}{\cellcolor[HTML]{66C194}93.09} &
  \multicolumn{1}{c|}{\cellcolor[HTML]{5CBD8E}97.35} \\ \cline{2-4} \cline{7-10} \cline{13-17} \cline{19-24} 
\multicolumn{1}{l}{\textbf{}} &
  Net-1 &
  Net-2 &
  Final &
   &
   &
  Net-1 &
  Net-2 &
  Net-3 &
  Final &
   &
   &
  Net-1 &
  Net-2 &
  Net-3 &
  Net-4 &
  Final &
   &
  Net-1 &
  Net-2 &
  Net-3 &
  Net-4 &
  Net-5 &
  Final
\end{tabular}}
\label{tab:per-class-mnist}
\end{subtable}
\bigskip
\begin{subtable}[t]{\textwidth}
\caption{}
\resizebox{\linewidth}{!}{%
\centering
\begin{tabular}{ccccllccccclccccclcccccc}
\multicolumn{1}{l}{} &
  \multicolumn{3}{c}{n = 2} &
   &
   &
  \multicolumn{4}{c}{n = 3} &
  \multicolumn{1}{l}{} &
   &
  \multicolumn{5}{c}{n = 4} &
   &
  \multicolumn{6}{c}{n = 5} \\ \cline{2-4} \cline{7-10} \cline{13-17} \cline{19-24} 
\multicolumn{1}{c|}{o} &
  \multicolumn{1}{c|}{\cellcolor[HTML]{95D5B6}76.26} &
  \multicolumn{1}{c|}{\cellcolor[HTML]{57BB8A}92.15} &
  \multicolumn{1}{c|}{\cellcolor[HTML]{63C092}92.61} &
   &
  \multicolumn{1}{l|}{} &
  \multicolumn{1}{c|}{\cellcolor[HTML]{98D5B7}66.90} &
  \multicolumn{1}{c|}{\cellcolor[HTML]{73C69D}77.87} &
  \multicolumn{1}{c|}{\cellcolor[HTML]{B7E2CD}57.51} &
  \multicolumn{1}{c|}{\cellcolor[HTML]{5FBF90}93.36} &
   &
  \multicolumn{1}{l|}{} &
  \multicolumn{1}{c|}{\cellcolor[HTML]{84CEAA}75.75} &
  \multicolumn{1}{c|}{\cellcolor[HTML]{66C195}86.90} &
  \multicolumn{1}{c|}{\cellcolor[HTML]{74C79E}81.82} &
  \multicolumn{1}{c|}{\cellcolor[HTML]{D1EDDF}47.54} &
  \multicolumn{1}{c|}{\cellcolor[HTML]{5DBE8F}93.47} &
  \multicolumn{1}{l|}{} &
  \multicolumn{1}{c|}{\cellcolor[HTML]{57BB8A}89.81} &
  \multicolumn{1}{c|}{\cellcolor[HTML]{58BC8B}89.66} &
  \multicolumn{1}{c|}{\cellcolor[HTML]{B1E0C9}57.80} &
  \multicolumn{1}{c|}{\cellcolor[HTML]{99D6B8}66.26} &
  \multicolumn{1}{c|}{\cellcolor[HTML]{D1EDDF}46.32} &
  \multicolumn{1}{c|}{\cellcolor[HTML]{5CBD8D}93.89} \\ \cline{2-4} \cline{7-10} \cline{13-17} \cline{19-24} 
\multicolumn{1}{c|}{ki} &
  \multicolumn{1}{c|}{\cellcolor[HTML]{9ED8BC}73.97} &
  \multicolumn{1}{c|}{\cellcolor[HTML]{76C8A0}84.29} &
  \multicolumn{1}{c|}{\cellcolor[HTML]{65C194}91.93} &
   &
  \multicolumn{1}{l|}{} &
  \multicolumn{1}{c|}{\cellcolor[HTML]{FFFFFF}36.16} &
  \multicolumn{1}{c|}{\cellcolor[HTML]{99D6B8}66.43} &
  \multicolumn{1}{c|}{\cellcolor[HTML]{A1D9BE}64.12} &
  \multicolumn{1}{c|}{\cellcolor[HTML]{66C195}90.91} &
   &
  \multicolumn{1}{l|}{} &
  \multicolumn{1}{c|}{\cellcolor[HTML]{CAEADA}50.00} &
  \multicolumn{1}{c|}{\cellcolor[HTML]{DAF0E5}44.32} &
  \multicolumn{1}{c|}{\cellcolor[HTML]{DEF2E8}42.90} &
  \multicolumn{1}{c|}{\cellcolor[HTML]{B3E0CA}58.70} &
  \multicolumn{1}{c|}{\cellcolor[HTML]{65C194}90.62} &
  \multicolumn{1}{l|}{} &
  \multicolumn{1}{c|}{\cellcolor[HTML]{F8FCFA}32.40} &
  \multicolumn{1}{c|}{\cellcolor[HTML]{D7EFE3}44.29} &
  \multicolumn{1}{c|}{\cellcolor[HTML]{DBF1E6}42.84} &
  \multicolumn{1}{c|}{\cellcolor[HTML]{ADDEC6}59.19} &
  \multicolumn{1}{c|}{\cellcolor[HTML]{F4FBF7}33.91} &
  \multicolumn{1}{c|}{\cellcolor[HTML]{62C091}91.58} \\ \cline{2-4} \cline{7-10} \cline{13-17} \cline{19-24} 
\multicolumn{1}{c|}{su} &
  \multicolumn{1}{c|}{\cellcolor[HTML]{FFFFFF}48.96} &
  \multicolumn{1}{c|}{\cellcolor[HTML]{E0F3EA}57.00} &
  \multicolumn{1}{c|}{\cellcolor[HTML]{78C9A1}86.77} &
   &
  \multicolumn{1}{l|}{} &
  \multicolumn{1}{c|}{\cellcolor[HTML]{A0D9BD}64.27} &
  \multicolumn{1}{c|}{\cellcolor[HTML]{A3DABF}63.46} &
  \multicolumn{1}{c|}{\cellcolor[HTML]{E7F6EF}43.32} &
  \multicolumn{1}{c|}{\cellcolor[HTML]{72C69D}86.82} &
   &
  \multicolumn{1}{l|}{} &
  \multicolumn{1}{c|}{\cellcolor[HTML]{FFFFFF}30.37} &
  \multicolumn{1}{c|}{\cellcolor[HTML]{DAF0E6}44.03} &
  \multicolumn{1}{c|}{\cellcolor[HTML]{CEECDD}48.54} &
  \multicolumn{1}{c|}{\cellcolor[HTML]{D2EDE0}46.96} &
  \multicolumn{1}{c|}{\cellcolor[HTML]{70C59B}86.40} &
  \multicolumn{1}{l|}{} &
  \multicolumn{1}{c|}{\cellcolor[HTML]{D5EEE2}44.73} &
  \multicolumn{1}{c|}{\cellcolor[HTML]{F1F9F5}35.02} &
  \multicolumn{1}{c|}{\cellcolor[HTML]{F1FAF6}34.72} &
  \multicolumn{1}{c|}{\cellcolor[HTML]{C7E9D8}49.94} &
  \multicolumn{1}{c|}{\cellcolor[HTML]{FFFFFF}29.66} &
  \multicolumn{1}{c|}{\cellcolor[HTML]{70C59C}85.88} \\ \cline{2-4} \cline{7-10} \cline{13-17} \cline{19-24} 
\multicolumn{1}{c|}{tsu} &
  \multicolumn{1}{c|}{\cellcolor[HTML]{C9EADA}62.89} &
  \multicolumn{1}{c|}{\cellcolor[HTML]{B5E1CB}68.20} &
  \multicolumn{1}{c|}{\cellcolor[HTML]{6BC398}90.31} &
   &
  \multicolumn{1}{l|}{} &
  \multicolumn{1}{c|}{\cellcolor[HTML]{65C194}81.94} &
  \multicolumn{1}{c|}{\cellcolor[HTML]{85CEAA}72.38} &
  \multicolumn{1}{c|}{\cellcolor[HTML]{6EC49A}79.35} &
  \multicolumn{1}{c|}{\cellcolor[HTML]{60BF91}92.96} &
   &
  \multicolumn{1}{l|}{} &
  \multicolumn{1}{c|}{\cellcolor[HTML]{B9E3CE}56.40} &
  \multicolumn{1}{c|}{\cellcolor[HTML]{CCEBDC}49.27} &
  \multicolumn{1}{c|}{\cellcolor[HTML]{B3E1CB}58.38} &
  \multicolumn{1}{c|}{\cellcolor[HTML]{88CFAC}74.32} &
  \multicolumn{1}{c|}{\cellcolor[HTML]{5DBE8F}93.52} &
  \multicolumn{1}{l|}{} &
  \multicolumn{1}{c|}{\cellcolor[HTML]{8FD2B1}70.08} &
  \multicolumn{1}{c|}{\cellcolor[HTML]{C3E7D5}51.31} &
  \multicolumn{1}{c|}{\cellcolor[HTML]{6FC59B}81.44} &
  \multicolumn{1}{c|}{\cellcolor[HTML]{6EC59A}81.71} &
  \multicolumn{1}{c|}{\cellcolor[HTML]{BDE4D1}53.56} &
  \multicolumn{1}{c|}{\cellcolor[HTML]{5EBE8F}92.85} \\ \cline{2-4} \cline{7-10} \cline{13-17} \cline{19-24} 
\multicolumn{1}{c|}{na} &
  \multicolumn{1}{c|}{\cellcolor[HTML]{B6E2CD}67.75} &
  \multicolumn{1}{c|}{\cellcolor[HTML]{D9F0E5}58.87} &
  \multicolumn{1}{c|}{\cellcolor[HTML]{70C69C}88.86} &
   &
  \multicolumn{1}{l|}{} &
  \multicolumn{1}{c|}{\cellcolor[HTML]{A9DCC3}61.84} &
  \multicolumn{1}{c|}{\cellcolor[HTML]{76C8A0}76.86} &
  \multicolumn{1}{c|}{\cellcolor[HTML]{CDEBDC}51.20} &
  \multicolumn{1}{c|}{\cellcolor[HTML]{66C195}90.96} &
   &
  \multicolumn{1}{l|}{} &
  \multicolumn{1}{c|}{\cellcolor[HTML]{BEE5D2}54.63} &
  \multicolumn{1}{c|}{\cellcolor[HTML]{7DCBA5}78.43} &
  \multicolumn{1}{c|}{\cellcolor[HTML]{D8F0E4}44.91} &
  \multicolumn{1}{c|}{\cellcolor[HTML]{8ED2B0}72.18} &
  \multicolumn{1}{c|}{\cellcolor[HTML]{65C194}90.45} &
  \multicolumn{1}{l|}{} &
  \multicolumn{1}{c|}{\cellcolor[HTML]{99D6B8}66.21} &
  \multicolumn{1}{c|}{\cellcolor[HTML]{C6E8D7}50.35} &
  \multicolumn{1}{c|}{\cellcolor[HTML]{E5F5ED}39.13} &
  \multicolumn{1}{c|}{\cellcolor[HTML]{CBEADB}48.31} &
  \multicolumn{1}{c|}{\cellcolor[HTML]{B9E3CF}54.84} &
  \multicolumn{1}{c|}{\cellcolor[HTML]{62C092}91.38} \\ \cline{2-4} \cline{7-10} \cline{13-17} \cline{19-24} 
\multicolumn{1}{c|}{ha} &
  \multicolumn{1}{c|}{\cellcolor[HTML]{8BD0AF}78.82} &
  \multicolumn{1}{c|}{\cellcolor[HTML]{72C69D}85.22} &
  \multicolumn{1}{c|}{\cellcolor[HTML]{57BB8A}95.81} &
   &
  \multicolumn{1}{l|}{} &
  \multicolumn{1}{c|}{\cellcolor[HTML]{8DD1B0}70.00} &
  \multicolumn{1}{c|}{\cellcolor[HTML]{57BB8A}85.87} &
  \multicolumn{1}{c|}{\cellcolor[HTML]{58BC8B}85.66} &
  \multicolumn{1}{c|}{\cellcolor[HTML]{57BB8A}96.11} &
   &
  \multicolumn{1}{l|}{} &
  \multicolumn{1}{c|}{\cellcolor[HTML]{C9E9DA}50.40} &
  \multicolumn{1}{c|}{\cellcolor[HTML]{AFDFC7}60.07} &
  \multicolumn{1}{c|}{\cellcolor[HTML]{57BB8A}92.27} &
  \multicolumn{1}{c|}{\cellcolor[HTML]{A5DBC1}63.64} &
  \multicolumn{1}{c|}{\cellcolor[HTML]{57BB8A}95.80} &
  \multicolumn{1}{l|}{} &
  \multicolumn{1}{c|}{\cellcolor[HTML]{C0E6D3}52.29} &
  \multicolumn{1}{c|}{\cellcolor[HTML]{B3E0CA}57.13} &
  \multicolumn{1}{c|}{\cellcolor[HTML]{6FC59B}81.25} &
  \multicolumn{1}{c|}{\cellcolor[HTML]{BCE4D0}53.85} &
  \multicolumn{1}{c|}{\cellcolor[HTML]{77C8A1}78.43} &
  \multicolumn{1}{c|}{\cellcolor[HTML]{57BB8A}95.51} \\ \cline{2-4} \cline{7-10} \cline{13-17} \cline{19-24} 
\multicolumn{1}{c|}{ma} &
  \multicolumn{1}{c|}{\cellcolor[HTML]{89CFAD}79.54} &
  \multicolumn{1}{c|}{\cellcolor[HTML]{A0D9BD}73.49} &
  \multicolumn{1}{c|}{\cellcolor[HTML]{72C69D}88.42} &
   &
  \multicolumn{1}{l|}{} &
  \multicolumn{1}{c|}{\cellcolor[HTML]{DBF1E6}47.10} &
  \multicolumn{1}{c|}{\cellcolor[HTML]{A9DDC3}61.70} &
  \multicolumn{1}{c|}{\cellcolor[HTML]{9CD7BA}65.72} &
  \multicolumn{1}{c|}{\cellcolor[HTML]{72C69D}86.67} &
   &
  \multicolumn{1}{l|}{} &
  \multicolumn{1}{c|}{\cellcolor[HTML]{7FCCA6}77.54} &
  \multicolumn{1}{c|}{\cellcolor[HTML]{DDF1E7}43.15} &
  \multicolumn{1}{c|}{\cellcolor[HTML]{96D5B6}69.14} &
  \multicolumn{1}{c|}{\cellcolor[HTML]{BBE4D0}55.65} &
  \multicolumn{1}{c|}{\cellcolor[HTML]{69C397}88.95} &
  \multicolumn{1}{l|}{} &
  \multicolumn{1}{c|}{\cellcolor[HTML]{E0F3E9}41.11} &
  \multicolumn{1}{c|}{\cellcolor[HTML]{B5E1CB}56.41} &
  \multicolumn{1}{c|}{\cellcolor[HTML]{CBEADB}48.51} &
  \multicolumn{1}{c|}{\cellcolor[HTML]{E1F3EA}40.66} &
  \multicolumn{1}{c|}{\cellcolor[HTML]{E9F6F0}37.87} &
  \multicolumn{1}{c|}{\cellcolor[HTML]{6BC498}87.75} \\ \cline{2-4} \cline{7-10} \cline{13-17} \cline{19-24} 
\multicolumn{1}{c|}{ya} &
  \multicolumn{1}{c|}{\cellcolor[HTML]{AEDFC7}69.87} &
  \multicolumn{1}{c|}{\cellcolor[HTML]{DFF2E9}57.35} &
  \multicolumn{1}{c|}{\cellcolor[HTML]{5EBE8F}93.95} &
   &
  \multicolumn{1}{l|}{} &
  \multicolumn{1}{c|}{\cellcolor[HTML]{70C59B}78.68} &
  \multicolumn{1}{c|}{\cellcolor[HTML]{71C69D}78.20} &
  \multicolumn{1}{c|}{\cellcolor[HTML]{70C59C}78.64} &
  \multicolumn{1}{c|}{\cellcolor[HTML]{5CBD8E}94.36} &
   &
  \multicolumn{1}{l|}{} &
  \multicolumn{1}{c|}{\cellcolor[HTML]{98D6B8}68.37} &
  \multicolumn{1}{c|}{\cellcolor[HTML]{B8E2CE}56.80} &
  \multicolumn{1}{c|}{\cellcolor[HTML]{94D4B4}70.08} &
  \multicolumn{1}{c|}{\cellcolor[HTML]{89D0AD}73.91} &
  \multicolumn{1}{c|}{\cellcolor[HTML]{5ABD8D}94.68} &
  \multicolumn{1}{l|}{} &
  \multicolumn{1}{c|}{\cellcolor[HTML]{B8E3CE}55.13} &
  \multicolumn{1}{c|}{\cellcolor[HTML]{BBE4D0}54.11} &
  \multicolumn{1}{c|}{\cellcolor[HTML]{AFDFC7}58.63} &
  \multicolumn{1}{c|}{\cellcolor[HTML]{8DD1B0}70.67} &
  \multicolumn{1}{c|}{\cellcolor[HTML]{B1E0C9}57.61} &
  \multicolumn{1}{c|}{\cellcolor[HTML]{5ABC8C}94.62} \\ \cline{2-4} \cline{7-10} \cline{13-17} \cline{19-24} 
\multicolumn{1}{c|}{re} &
  \multicolumn{1}{c|}{\cellcolor[HTML]{A4DAC0}72.54} &
  \multicolumn{1}{c|}{\cellcolor[HTML]{7AC9A2}83.40} &
  \multicolumn{1}{c|}{\cellcolor[HTML]{6AC397}90.61} &
   &
  \multicolumn{1}{l|}{} &
  \multicolumn{1}{c|}{\cellcolor[HTML]{6CC499}79.91} &
  \multicolumn{1}{c|}{\cellcolor[HTML]{B3E0CA}58.84} &
  \multicolumn{1}{c|}{\cellcolor[HTML]{A2DABF}63.71} &
  \multicolumn{1}{c|}{\cellcolor[HTML]{66C195}90.89} &
   &
  \multicolumn{1}{l|}{} &
  \multicolumn{1}{c|}{\cellcolor[HTML]{E5F5ED}40.09} &
  \multicolumn{1}{c|}{\cellcolor[HTML]{7DCBA5}78.33} &
  \multicolumn{1}{c|}{\cellcolor[HTML]{94D4B4}70.06} &
  \multicolumn{1}{c|}{\cellcolor[HTML]{DAF0E5}44.14} &
  \multicolumn{1}{c|}{\cellcolor[HTML]{67C295}89.86} &
  \multicolumn{1}{l|}{} &
  \multicolumn{1}{c|}{\cellcolor[HTML]{D4EEE1}45.38} &
  \multicolumn{1}{c|}{\cellcolor[HTML]{C8E9D9}49.66} &
  \multicolumn{1}{c|}{\cellcolor[HTML]{DBF1E6}42.62} &
  \multicolumn{1}{c|}{\cellcolor[HTML]{D9F0E4}43.62} &
  \multicolumn{1}{c|}{\cellcolor[HTML]{C9EADA}49.06} &
  \multicolumn{1}{c|}{\cellcolor[HTML]{63C093}91.00} \\ \cline{2-4} \cline{7-10} \cline{13-17} \cline{19-24} 
\multicolumn{1}{c|}{wo} &
  \multicolumn{1}{c|}{\cellcolor[HTML]{B5E1CB}68.23} &
  \multicolumn{1}{c|}{\cellcolor[HTML]{A8DCC2}71.51} &
  \multicolumn{1}{c|}{\cellcolor[HTML]{5EBE8F}94.10} &
   &
  \multicolumn{1}{l|}{} &
  \multicolumn{1}{c|}{\cellcolor[HTML]{78C9A1}76.15} &
  \multicolumn{1}{c|}{\cellcolor[HTML]{F0F9F5}40.68} &
  \multicolumn{1}{c|}{\cellcolor[HTML]{BBE4D0}56.52} &
  \multicolumn{1}{c|}{\cellcolor[HTML]{61BF91}92.71} &
   &
  \multicolumn{1}{l|}{} &
  \multicolumn{1}{c|}{\cellcolor[HTML]{A0D9BD}65.73} &
  \multicolumn{1}{c|}{\cellcolor[HTML]{76C89F}81.18} &
  \multicolumn{1}{c|}{\cellcolor[HTML]{EEF8F3}36.86} &
  \multicolumn{1}{c|}{\cellcolor[HTML]{A5DBC1}63.54} &
  \multicolumn{1}{c|}{\cellcolor[HTML]{5FBF90}92.88} &
  \multicolumn{1}{l|}{} &
  \multicolumn{1}{c|}{\cellcolor[HTML]{B0DFC8}58.00} &
  \multicolumn{1}{c|}{\cellcolor[HTML]{81CCA7}75.00} &
  \multicolumn{1}{c|}{\cellcolor[HTML]{B8E3CE}55.14} &
  \multicolumn{1}{c|}{\cellcolor[HTML]{CAEADA}48.84} &
  \multicolumn{1}{c|}{\cellcolor[HTML]{93D4B4}68.35} &
  \multicolumn{1}{c|}{\cellcolor[HTML]{5BBD8D}94.15} \\ \cline{2-4} \cline{7-10} \cline{13-17} \cline{19-24} 
\textbf{} &
  Net-1 &
  Net-2 &
  Final &
   &
   &
  Net-1 &
  Net-2 &
  Net-3 &
  Final &
   &
   &
  Net-1 &
  Net-2 &
  Net-3 &
  Net-4 &
  Final &
   &
  Net-1 &
  Net-2 &
  Net-3 &
  Net-4 &
  Net-5 &
  Final
\end{tabular}}
\label{tab:per-class-kmnist}
\end{subtable}
\label{tab:per-class-acc}
\end{table}
\indent kWTA-ENN uses a kWTA activation function so that the neurons from its sub-networks could compete for their right to respond to a particular subset of the dataset. We demonstrate this in Figures \ref{fig:mnist-per-class-logits} and \ref{fig:kmnist-per-class-logits}. Let us recall that kWTA-ENN gets its outputs by computing a linear combination of the outputs of its sub-networks, and then passing the linear combination results to a kWTA activation function. As per the referred figures, even though each kWTA-ENN sub-network is not providing high probability output per class as compared to MoE and CE sub-networks, the final kWTA-ENN output is on par with the MoE and CE probability outputs.\\
\indent We can then infer two things from this: (1) the kWTA activation function inhibits the neurons of the losing kWTA-ENN sub-networks, and (2) the probability outputs of the winning sub-network neurons enable the sub-networks to help one another. For instance, in Figure \ref{fig:kwta-mnist-per-class-logits}, we can observe a probability output for class 1 from sub-network 1, however minimal, and a higher probability output for class 1 from sub-networks 2 and 3, but then their final output has even higher probability output for the same class when compared to MoE and CE probability outputs. The same could be observed in Figure \ref{fig:kwta-kmnist-per-class-logits}. This is because the losing neurons are inhibited in the competition process while retaining the winner neurons, thus improving the final probability output of the model.\\
\indent In Table \ref{tab:per-class-acc}, we further support this by showing the per-class accuracy of each kWTA-ENN sub-network with varying number of sub-networks. We can see that there is some apparent division of classes among the sub-networks even without pre-defining such divisions, but the final per-class accuracies of the entire model are even better than the per-class accuracies of the sub-networks, thus suggesting that there is indeed a sharing of responsibility among the sub-networks due to the inhibition of losing sub-network neurons and retention of the winning sub-network neurons, even with the competition in place.

\section{Conclusion and Future Works}
We introduce the k-Winners-Take-All ensemble neural network (kWTA-ENN) which uses a kWTA activation function as the means to combine the sub-network outputs in an ensemble as opposed to the conventional way of combining sub-network outputs through averaging, summation, or voting. Using a kWTA activation function induces competition among the sub-network neurons in an ensemble. This in turn leads to some form of specialization among them, thereby improving the overall performance of the ensemble.\\
\indent Our comparative results showed that our proposed approach outperforms our baseline models, yielding the following test accuracies on benchmark datasets: 98.34\% on MNIST, 88.06\% on Fashion-MNIST, 91.56\% on KMNIST, and 95.97\% on WDBC. We intend to pursue further exploration into this subject by comparing the performance of our baseline and experimental models with respect to varying mini-batch sizes, by training on other benchmark datasets, and finally, by using a more rigorous statistical treatment for a more formal comparison between our proposed model and our baseline models.

\bibliographystyle{splncs04}
\bibliography{mybibliography}

\end{document}